\documentclass[10pt,twocolumn,letterpaper]{article}

\usepackage[]{cvpr}  
\usepackage[ruled,linesnumbered]{algorithm2e}
\usepackage{algorithmic}
\usepackage{multirow}
\usepackage{booktabs}
\usepackage{wrapfig}
\usepackage{enumitem}

\usepackage[dvipsnames]{xcolor}

\definecolor{cvprblue}{rgb}{0.21,0.49,0.74}
\usepackage[pagebackref,breaklinks,colorlinks,citecolor=cvprblue]{hyperref}
\usepackage{graphicx}

\def\confName{CVPR}
\def\confYear{2024}
\begin{document}

\title{Generating Daylight-driven Architectural Design via Diffusion Models}

\author{Pengzhi Li\\
Tsinghua University\\
{\tt\small lpz21@mails.tsinghua.edu.cn}
\and
Baijuan Li\\
SAUP, Shenzhen University\\
{\tt\small 2060321010@email.szu.edu.cn}
}

\twocolumn[{%
\maketitle
\renewcommand\twocolumn[1][]{#1}%
\begin{center}
    \centering

  \includegraphics[width=1.0\textwidth]{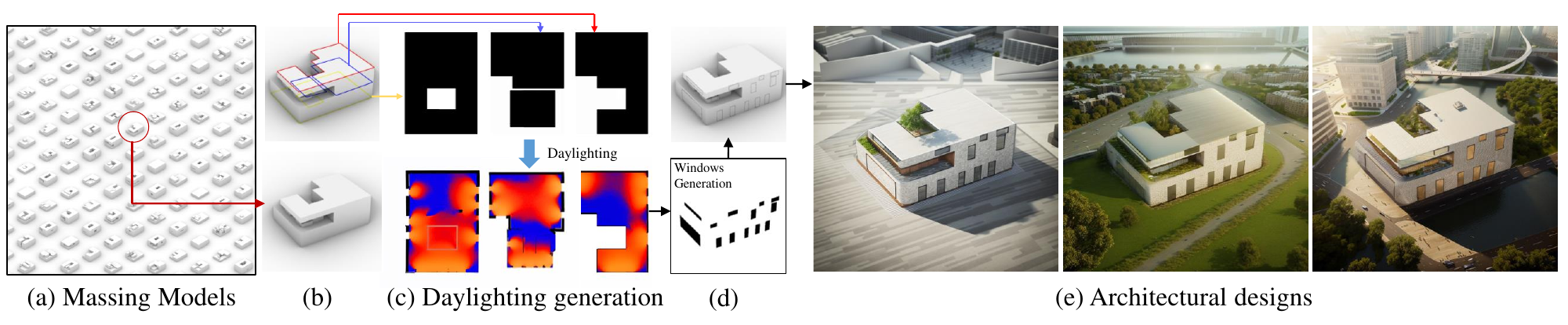}

      \captionsetup{type=figure}\caption{
    We propose a novel AI-aided architectural design method. We first generate numerous massing models as shown in (a). Then, a specific massing model (b) is selected. Following this, guided by a daylight-driven strategy, we refine the architectural façade depicted in (d), achieving efficient generations of the architectural design in (e) by using diffusion models.
    }
    \label{fig:teaser}
\end{center}
}]


\begin{abstract}

In recent years, the rapid development of large-scale models has made new possibilities for interdisciplinary fields such as architecture. In this paper, we present a novel daylight-driven AI-aided architectural design method. Firstly, we formulate a method for generating massing models, producing architectural massing models using random parameters quickly. Subsequently, we integrate a daylight-driven façade design strategy, accurately determining window layouts and applying them to the massing models. Finally, we seamlessly combine a large-scale language model with a text-to-image model, enhancing the efficiency of generating visual architectural design renderings. Experimental results demonstrate that our approach supports architects' creative inspirations and pioneers novel avenues for architectural design development. Project page: \textcolor{magenta}{https://zrealli.github.io/DDADesign/}.

\end{abstract}

\section{Introduction}
As an interdisciplinary field encompassing engineering and art, architecture blends elements of technology and creativity. The distinctive interdisciplinary adds complexity to the architectural design process~\cite{granadeiro2013building,hosseini2019morphologicaldfacade}, necessitating the simultaneous consideration of aesthetic design and functional layout. This complexity results in the initial stages of architectural design demanding significant time and effort. Traditionally, architects commence with fundamental floor sketches, gradually evolving them into massing models, often relying on manual work within 3D modeling software. However, this process inevitably involves numerous time-consuming revisions and adjustments.

In recent years, with the advancement of deep learning techniques, several studies have introduced methods capable of autonomously generating architectural floorplans~\cite{sun2022wallplan,shabani2023housediffusion,nauata2021house}. These methods can automatically produce functional layouts based on the outlines of floorplans. However, these approaches primarily focus on developing floorplans for residential buildings while neglecting crucial architectural factors such as daylighting~\cite{kischkoweit2002overview,wong2017review}. Furthermore, they cannot encompass the entirety of the architectural design process. Recently, significant progress has been made in large-scale text-to-image models~\cite{rombach2022stable,saharia2022photorealistic,saharia2022imagen,ramesh2022hierarchical,nichol2021glide}. These models exhibit potent content generation abilities and exceptional aesthetic expression skills, presenting novel opportunities for reimagining design~\cite{li2023sketch,makatura2023can,liao2021automated,chaillou2020archigan,para2021generative}. Leveraging text prompts, we can generate entirely new architectural designs from sketches. This approach holds immense potential while also being a challenging and fascinating task.

In this paper, we propose a novel AI-aided architectural design method to generate preliminary architectural designs from massing models. Unlike S2A~\cite{li2023sketch}, which relies on using predicted depth map~\cite{Ranftl2022} to generate masssing models, it's difficult to obtain consistent and accurate edge depth maps~\cite{li2023efficient,li2023towards,li2024devil} from single-frame perspective images. Therefore, we utilize 3D modeling software to achieve more precise massing generation.
We first introduce a method for generating massing models, drawing inspiration from traditional architectural design strategies that involve iteratively adding and subtracting volumes to refine the design. We visualize this strategy within Grasshopper and, by adjusting random parameters, are able to generate a diverse array of massing models swiftly.
Subsequently, we present a façade generation approach centered around daylighting. Given the lack of readily available daylighting data, we undertake a series of steps to construct a dedicated dataset of daylighting maps for training the LoRA (Low-Rank Adaptation) model~\cite{hu2021lora}. We then input representative sectional profiles of massing models into the trained model to generate corresponding daylighting maps. This process accurately determines window layouts and precisely integrates them onto the massing models.

Finally, we categorize key architectural terms into four major groups and employ the advanced language model GPT-4~\cite{openai2023gpt4} to generate corresponding text prompts randomly. These prompts are then fed into the Stable Diffusion v1.5 model~\cite{rombach2022stable}. To ensure alignment between the generated architectural designs and the contours of the massing models, we integrate additional and conditional controls into the diffusion model using ControlNet~\cite{zhang2023adding}. This enhancement aims to attain more precise outcomes in an architectural generation. Experimental results show that our approach efficiently assists architects in swiftly conceptualizing and creating initial design proposals. This paves a new pathway for the advancement of the architectural design. We summarize our main contributions as follows:

\begin{itemize}
\item {\verb||}We present a novel AI-aided architectural design method driven by daylight, enabling an end-to-end generation from massing models to architectural designs. This process offers support for architects' creative inspirations.
\item   We are the first to incorporate daylight factors into deep learning-based architectural generation methods and present a strategy for constructing an architectural daylighting dataset.
\item We seamlessly integrate large-scale language models with text-to-image models, enhancing the efficiency of generating visual architectural design renderings.
\end{itemize} 

\section{Method}
\begin{figure}[t]
    \centering
    \includegraphics[width=0.47\textwidth]{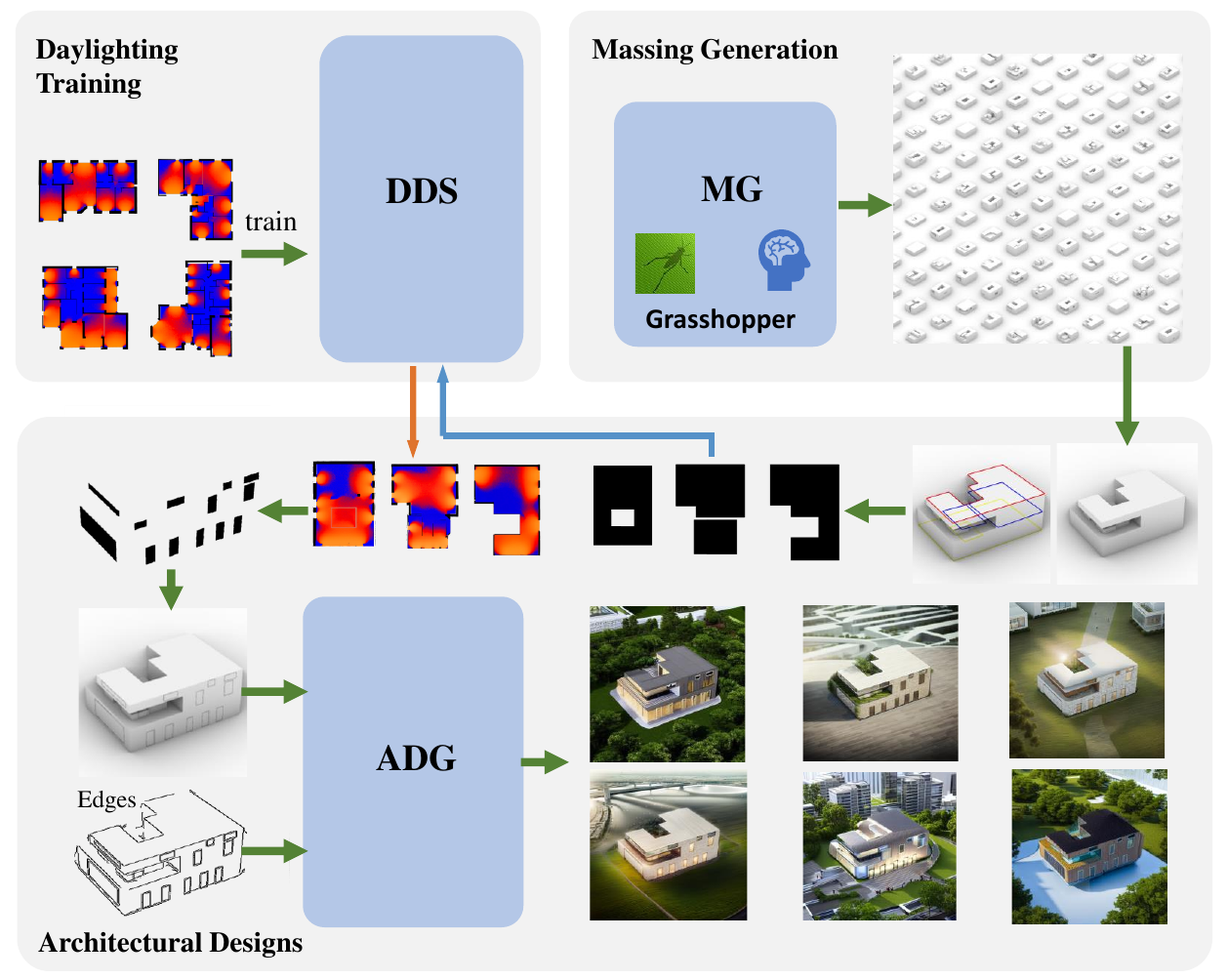}   

    \caption{Our pipeline shows an AI-aided architectural design workflow. Massing Generation (MG) is the module for generating architectural massing models, DDS is the daylight-driven strategy, and Architectural Design (ADG) is the module utilizing large-scale models to generate architectural designs. }
\vspace{-1em}
    \label{fig:pipeline}
\end{figure}

As shown in Fig.~\ref{fig:pipeline}, we describe our architectural design pipeline. Next, we describe each part in detail.

\subsection{Massing model generation}

\begin{figure}[t]
    \centering
    \includegraphics[width=0.43\textwidth]{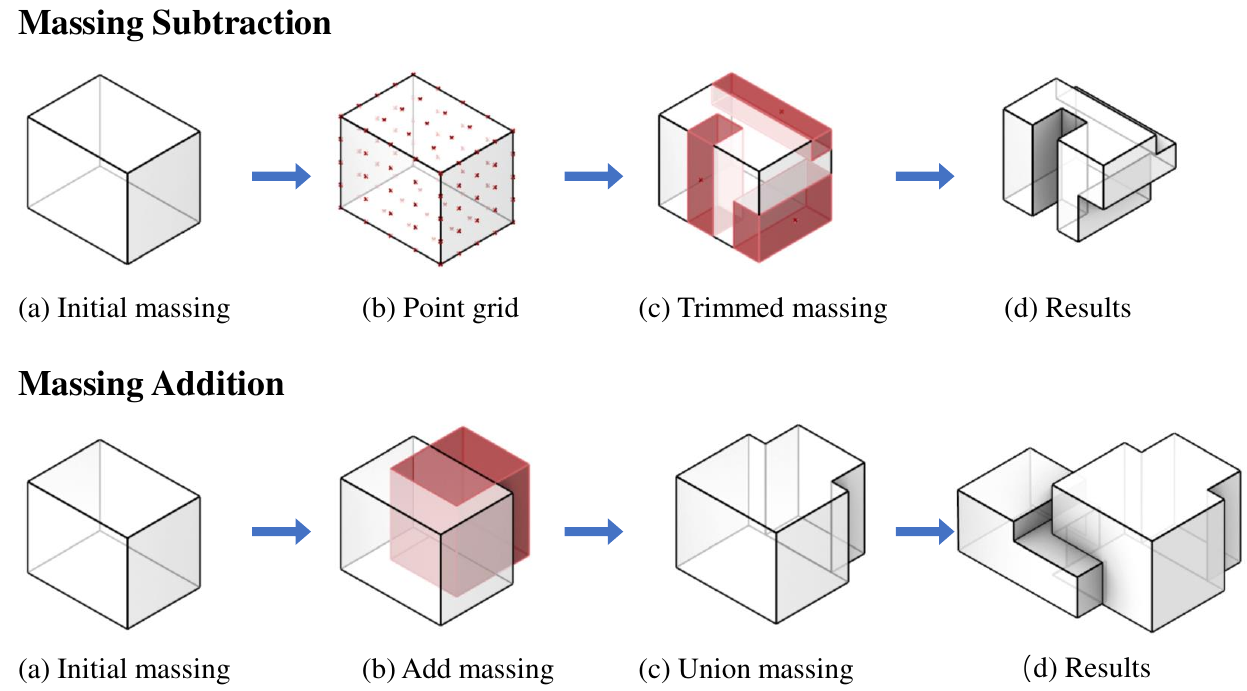}   
    \caption{We demonstrate the processes of addition and subtraction in the generation of massing models. The generation of new massing models is founded upon these fundamental operations.}

    \label{fig:massing generation}
 
\end{figure}

In the initial stages of architectural design ideation, designers commonly utilize massing models to refine the architectural form. This process involves techniques like massing addition and subtraction. We draw inspiration from these widely employed spatial design methods in architecture and formulate a parametric algorithm. As shown in Fig.~\ref{fig:massing generation}, our algorithm comprises two key components, focusing on manipulating massing by adding and subtracting massings.

We first establish an initial cubic massing model for the approach focused on massing addition. We implement additive modifications to this massing by employing translation and scaling actions. Subsequently, a Boolean union operation is conducted on all the massings, resulting in the initial massings used for architectural generation. Conversely, we define a cubic space for the massing subtraction approach where all three dimensions (x, y, z) are set to a maximum value according to the initial massing. Within this space, the generation of cubes for the 'massing subtraction' is guided by specific parameters. To enhance precise control, we ensure that the initial points used for massing generation align precisely with the edges. We use the Rhino-grasshopper platform~\cite{mcneel-rhinoceros} to implement our algorithm to facilitate parameter control through a visual graphical user interface. More comprehensive understanding and specifics regarding parameter control can be found in the supplementary materails.

\begin{figure}[t]
    \centering
    \includegraphics[width=0.37\textwidth]{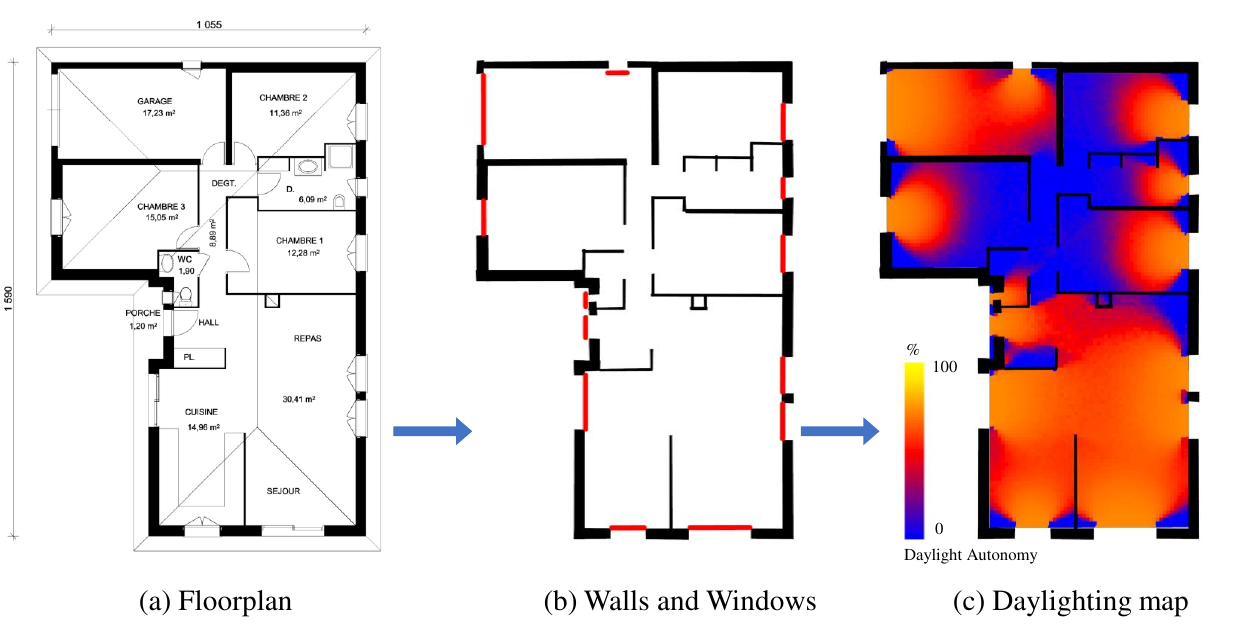}   
    \caption{We illustrate the process of generating our daylighting data. First, the floorplan (a) is decomposed into three vector components (b): interior walls, exterior walls, and windows (red). Subsequently, daylight parameters and algorithms are employed to compute the corresponding daylighting maps.}
   
    \label{fig:dataset}
 \vspace{-1em}
\end{figure}

\subsection{Daylight-driven strategy}

Architectural façade design~\cite{lim2012buildingfacade,moghtadernejad2019facade,hosseini2019morphologicaldfacade,sun2022automatic} is a crucial step in the process of architectural design. As a significant factor influencing façade design, daylight has a notable impact on indoor comfort and energy efficiency. However, previous researches focus on the automated generation of architectural floorplans~\cite{nauata2021house,sun2022wallplan,shabani2023housediffusion,hu2020graph2plan,para2021genarch}, and most efforts have concentrated solely on rational spatial layout, neglecting the importance of daylighting in architecture. To address this issue, we introduce a daylight-driven approach to architectural façade design, aiming to effectively utilize daylighting data for optimizing façade design. 

However, to the best of our knowledge, there is currently no existing daylighting dataset. Considering the rapid advancement of diffusion models in image-related tasks~\cite{couairon2022diffedit,li2023layerdiffusion,li2023sketch,brooks2023instructpix2pix,avrahami2023blended,ruiz2022dreambooth} and the introduction of LoRA~\cite{hu2021lora} technology, which eliminates the necessity for extensive training data, only a small amount of data is required for fine-tuning diffusion models to achieve the generation of similar image content. Therefore, we collect approximately 100 floorplans of various types of buildings. As shown in Fig.~\ref{fig:dataset}, these floorplans are further disassembled into vectorized drawings of windows, exterior walls, and interior walls. Subsequently, we employ the solar irradiation parameters specific to the Guangdong province within the Grasshopper software to conduct daylight analysis on these floorplans. This process produces plausible daylighting maps. 
Additionally, in the supplementary materials, we explains our devised daylighting map generation procedure and approach, realized through a plugin named ‘Honeybee’ within the Grasshopper software.
These daylighting maps assist the model in learning the distribution of window placements on building façades.

Subsequently, we employ the Stable Diffusion v1.5 model~\cite{rombach2022stable}, utilizing the generated daylighting map data as training input. Through integrating LoRA~\cite{hu2021lora} technology for model fine-tuning, we successfully obtain a precise LoRA~\cite{hu2021lora} model tailored for developing rational daylighting maps. As illustrated in Fig.~\ref{fig:pipeline}, we input the sectional profiles of each floor level of the architectural massing model and employ ControlNet~\cite{zhang2023adding} to control the plane outlines, culminating in producing corresponding daylight analysis visuals. Subsequently, utilizing these daylighting maps as references, we can accurately draw the window placements on the architectural massing. This daylight-driven design approach not only assists architects in completing façade designs but also ensures the rationality of indoor daylighting, which the previous studies overlook.

\begin{figure*}[t]
    \centering
    \includegraphics[width=0.98\textwidth]{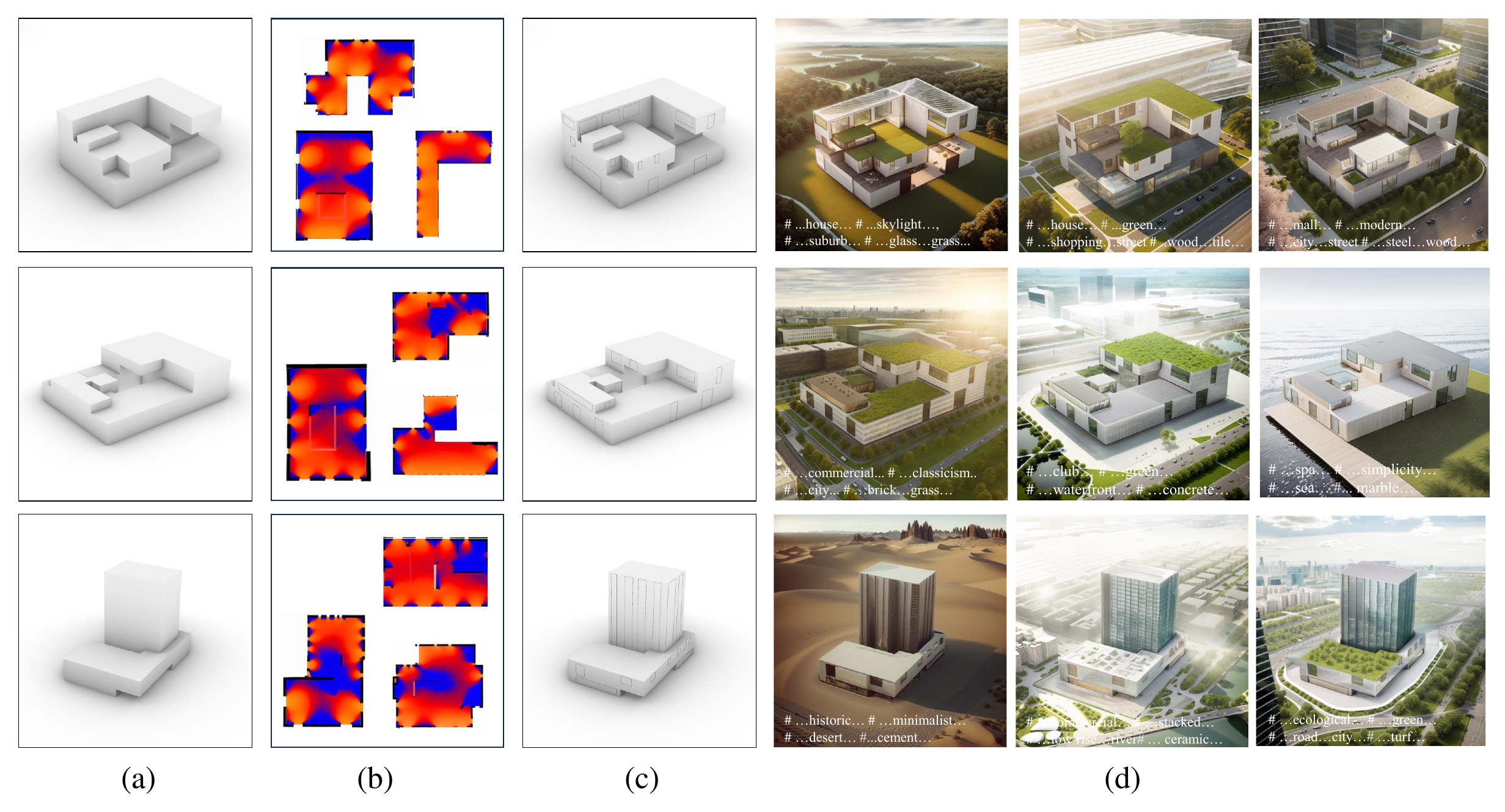}   
    \caption{More architectural designs. (a) is the generated massing model, (b) illustrates the daylighting map corresponding to (a). (c) shows the massing model following façade optimization, and (d) displays the visual renderings of the architectural design. '\#' divides the text prompts into four corresponding categories.}
       \vspace{-1em}
    \label{fig:visualization1}
 
\end{figure*}

\subsection{Architectural design generation}
We represent the diffusion process of a pre-trained model~\cite{rombach2022stable} as follows:

\begin{equation}\label{e:Phi_series}
\begin{array}{lcr}
     I_T, I_{T-1}, \ldots ,I_0. \ I_{t-1}=D(I_t|y),
\end{array}
\end{equation}
the $D$ is an update process : $\mathcal{I} \times \mathcal{C} \rightarrow \mathcal{I}$, $\mathcal{I} \in \mathbb{R}^{H \times W \times C}$ represents the image space, and $\mathcal{C}$ is the condition space, and $y \in \mathcal{C}$ is a text prompt. From ${T}$ to ${0}$, $I_{T}$ gradually changes from a Gaussian noise distribution being transformed into the desired image by $y$.

The role of text prompts $y$ is crucial during the process of architectural generation. However, for architectural generation, the core vocabulary encompasses many essential terms~\cite{ching2011visualarchi}. We categorize them into four categories: architectural types, architectural styles, architectural landscapes, and architectural materials. Recent developments in large-scale language models, such as GPT-4~\cite{openai2023gpt4}, which have been trained on extensive multimodal datasets, have achieved remarkable breakthroughs. To achieve text-controlled architectural generation, within GPT-4~\cite{openai2023gpt4}, by inputting instructions like "Generate N architectural design prompts, including architectural types, architectural styles, architectural landscape...." we can access a multitude of phrases for use in architectural generation.
Let A, B, C, and D denote the type, style, landscape, and material, respectively. The architectural design component $ADG$ can be defined as:
\begin{equation}\label{e:Pes}
\begin{array}{lcr}
     \{P_0, P_1, \ldots ,P_K.\} = ADG(A, B, C, D, K) 
\end{array}
\end{equation}
where $K$ is the number of prompts to generate. $P_i$ is the i-th prompt.

Furthermore, to maintain the generated results close to the appearance of the massing model, we use ControlNet~\cite{zhang2023adding} to add more conditions in the stable diffusion model~\cite{rombach2022stable} to control the precise architectural designs.

\section{Experimental Results}

We utilize random parameters to generate a series of massing models, employing them as the foundational elements for architectural design. As illustrated in Fig.~\ref{fig:visualization1}, our approach involves the initial extraction of three representative sectional profiles based on contour lines. Subsequently, we generate corresponding daylighting maps using the trained LoRA~\cite{hu2021lora} model. Due to the stochastic nature of the diffusion model, we meticulously filter the generated daylighting maps, ensuring the emergence of more plausible architectural façade designs. Based on the filtered daylighting maps, we create façade window profiles in (c). We generate many distinctive architectural design proposals in the final stages, leveraging diverse architectural text prompts. A subset of these results is showcased in (d), where these design proposals demonstrate remarkable performance concerning exterior proportions and visual renderings. Our approach is a valuable tool for architects to expedite the realization of architectural concepts and creativity, providing robust support for their creative endeavors.

\begin{table}[t]
  \centering
  \caption{User studies show that our framework is effective.}
  \label{table:comparison_user}
  \resizebox{.96\columnwidth}{!}{
    \begin{tabular}{lccc}
      \toprule
      Questions & Positive  & Negative  & Neutral \\   
      \midrule
      Daylighting strategy? & \textbf{64\%}  & 22\%  & 14\% \\
      Design efficiency?  & \textbf{88\%}  & 4\%  & 8\% \\
      Design inspiration? & \textbf{84\%}  & 4\%  & 12\% \\
      Design rationality? & \textbf{72\%}  & 8\%  & 20\% \\
      \bottomrule
    \end{tabular}
  }
\end{table}

\paragraph{User study:} Since our method aims to improve the design efficiency of architects, we survey 50 architects from different backgrounds about the improvements our method brings to practical design. We set several questions, such as whether the daylighting generation strategy is reasonable, whether the design efficiency is improved, whether design inspiration, the rationality of architectural design, and whether they are willing to use it. As shown in Tab.~\ref{table:comparison_user}, our method is recognized by most professional architects, further verifying the practical applicability of our approach. 

\section{Future work}

In the future, we aim to extend our design strategies to encompass more architectural physical factors and validate them across a more comprehensive range of architectural typologies. Enhancing the model's generation stability is a pivotal direction for future research. We are eagerly anticipating the emergence of more innovative achievements within architectural design.

\section{Conclusion}

This paper makes significant contributions to the exploration and advancement of architectural design. By employing a massing model generation approach and introducing an innovative daylight-driven AI-aided architectural design method, we successfully generate from massing models to preliminary architectural designs. Furthermore, we combine large-scale language models with text-to-image models efficiently produce architectural renderings. We firmly believe that this study provides crucial guidance and insight for future architectural design development.

\section{Supplementary materials}

In the supplementary materials, we primarily provide additional technical details to augment the content of the main text. Firstly, we showcase a selection of datasets mentioned in the section regarding daylight-driven strategies in Figure 6. This dataset includes floorplans and computationally derived daylighting maps. Subsequently, we illustrate various control parameters employed during the massing model generation process in Figure 7. Additionally, Figure 8 explains our devised daylighting map generation procedure and approach, realized through a plugin named ‘Honeybee’ within the Grasshopper software.

{
    \small
    \bibliographystyle{ieeenat_fullname}
    \bibliography{main}
}

\begin{figure*}
    \centering
    \includegraphics[width=0.95\textwidth]{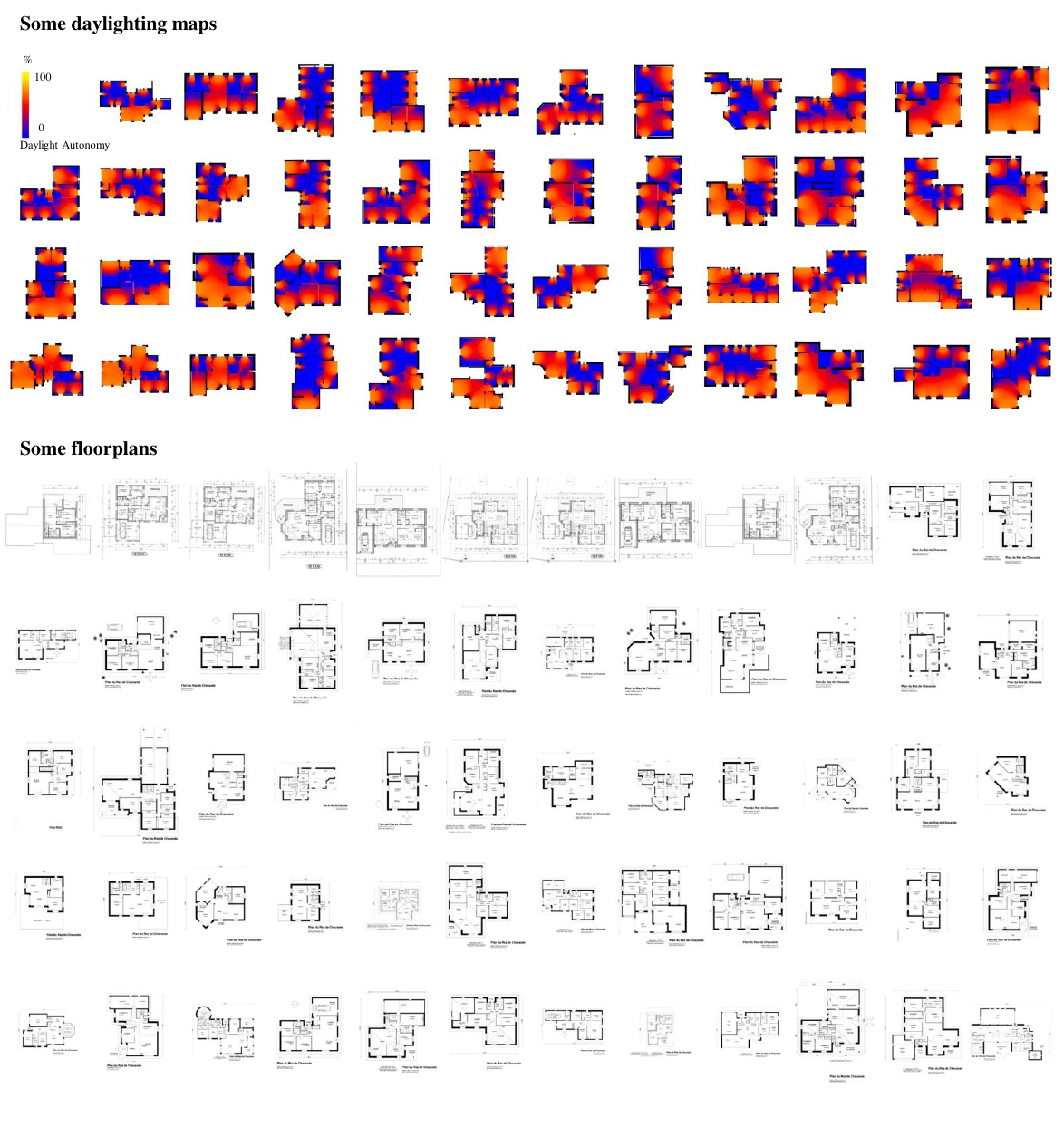}

    \vspace{-1em}
    \caption{We present more results of daylighting data generation. The upper figure showcases partial results of daylighting maps, while the lower figure shows some collected floorplans. }
    \label{fig:supp6}
 
\end{figure*}
\clearpage

\begin{figure*}[h]
    \centering
    \includegraphics[width=0.97\textwidth]{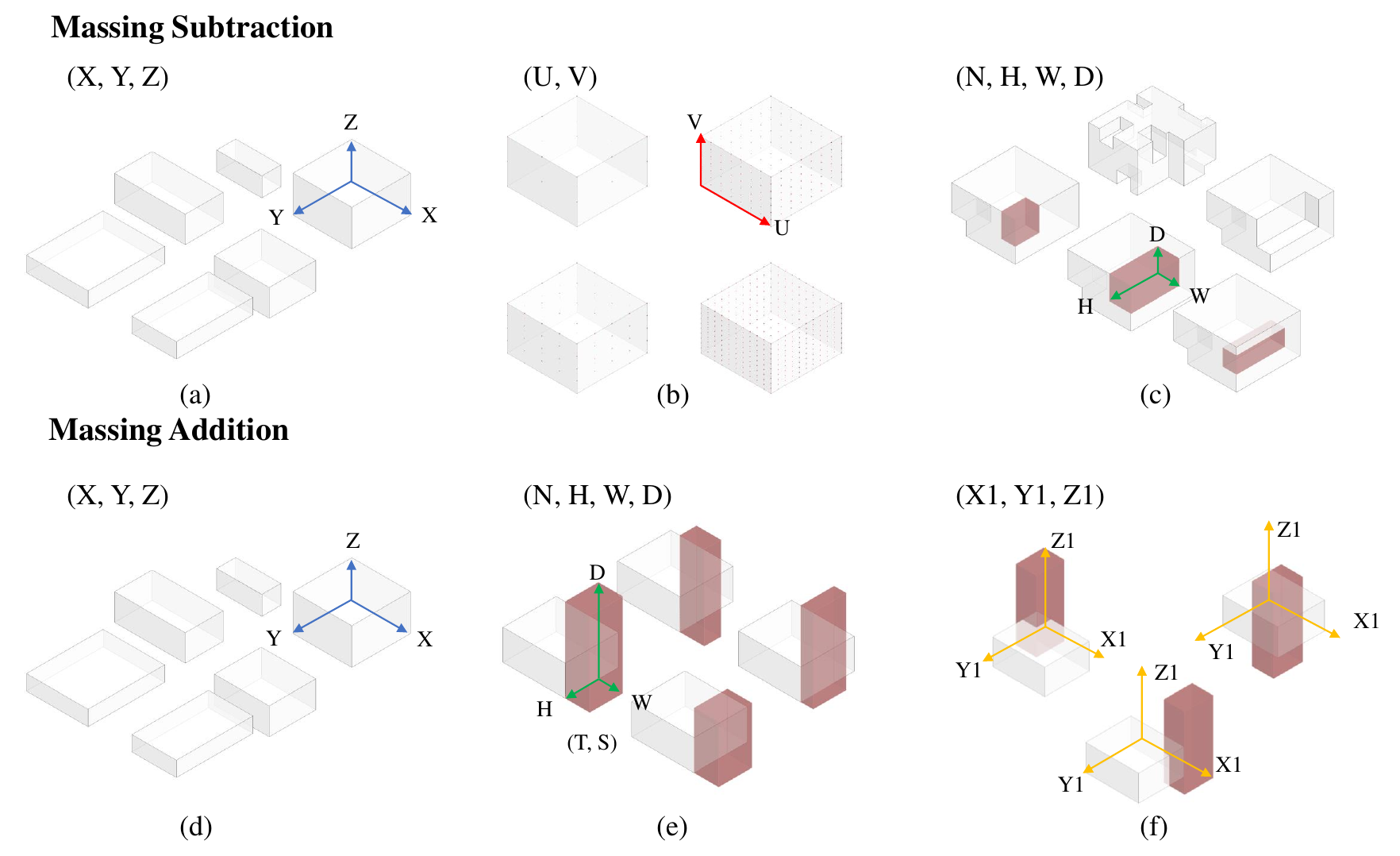}   

    \vspace{-1em}
    \caption{In step (a), we initiate the process by defining a cube with fixed dimensions XYZ. Subsequently, in step (b), we perform a surface segmentation on this cube along the UV directions. This segmentation generates a mesh composed of points in the UV coordinate system. During the subsequent massing block generation phase, we randomly select these points as the central points for generating the massing blocks. Then, in step (c), we utilize the parameter N to determine the number of subtraction operations to be executed on the massing. Additionally, we employ the parameters D, H, and W to individually control the dimensions used for partitioning the massings' height, length, and width. Adjusting these parameters allows us to manage the entirety of the massing subtraction process precisely. Like step (a), for massing addition, we proceed with step (d). Subsequently, in step (e), we similarly employ the parameters (N, H, W, D) to regulate the creation of massing blocks intended for addition. These blocks are positioned along the outline at the base of the cube. Following this, we apply translation (T) and scaling (S) operations to establish a new coordinate system (X1, Y1, Z1). This new system effectively encapsulates each newly generated massing's size throughout the steps. These procedures allow us to control massing subtraction and addition precisely.}
       \vspace{1em}
    \label{fig:supp99}
 
\end{figure*}

\begin{figure*}[h]
    \centering
    \includegraphics[width=0.9\textwidth]{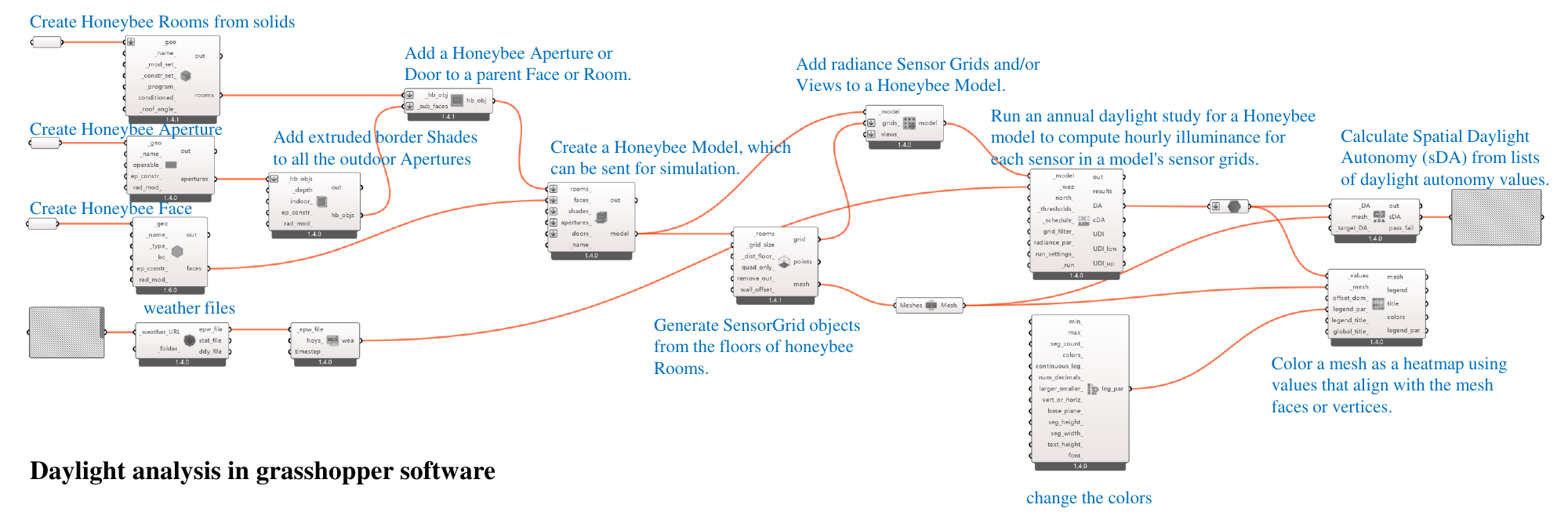}   

    \caption{We employ a visual programming approach within Grasshopper to comprehensively demonstrate our daylighting data generation methodology. The blue text labels above each box signify instructions for each respective operation.  }
    \label{fig:supp90}
 
\end{figure*}


\end{document}